\documentclass[11pt,letterpaper]{article}
\usepackage{emnlp2017}

\emnlpfinalcopy

\def\emnlppaperid{XXX}

\usepackage{mathptmx}

\usepackage{microtype}

\usepackage{comment}

\usepackage{enumitem}
\setlist{nolistsep}

\usepackage{subfig}

\usepackage{lipsum}
\usepackage{color}

\usepackage{graphicx}

\def\parsum#1{\bgroup {\textcolor{cyan}{\textbf{Paragraph summary:} #1}}\egroup}
\def\sectionsum#1{\bgroup \textcolor{green}{\textbf{Section content:} #1}\egroup \\}
\def\sectionsum#1{}
\def\parsum#1{}

\def\todo#1{\bgroup \textcolor{red}{\textbf{TODO:} #1} \egroup}

\title{Argotario: Computational Argumentation Meets Serious Games}

\author{Ivan Habernal,
Raffael Hannemann,
Christian Pollak, \\
\textbf{Christopher Klamm,}
\textbf{Patrick Pauli,}
\textbf{Iryna Gurevych}\\[.3em]
Ubiquitous Knowledge Processing Lab (UKP)\\
	Department of Computer Science, Technische Universit\"{a}t Darmstadt\\
	{\tt www.ukp.tu-darmstadt.de}\\
}

\date{}

\begin{document}
\maketitle
\begin{abstract}

An important skill in critical thinking and argumentation is the ability to spot and recognize \emph{fallacies}. Fallacious arguments, omnipresent in argumentative discourse, can be deceptive, manipulative, or simply leading to `wrong moves' in a discussion. 
Despite their importance, argumentation scholars and NLP researchers with focus on argumentation quality have not yet investigated fallacies empirically.
The nonexistence of resources dealing with fallacious argumentation calls for scalable approaches to data acquisition and annotation, for which the \emph{serious games} methodology offers an appealing, yet unexplored, alternative.
We present \emph{Argotario}, a serious game that deals with fallacies in everyday argumentation.
\emph{Argotario} is a multilingual, open-source, platform-independent application with strong educational aspects, accessible at \url{www.argotario.net}.
\end{abstract}

\section{Introduction}

\parsum{Computational argumentation (argument mining): not only structure but also quality issues}
Argumentation in natural language has been gaining much interest in the NLP community in recent years. While understanding the structure of an argument is the predominant task of argument mining/computational argumentation \cite{Mochales.2011,Stab.Gurevych.2014.EMNLP,Habernal.Gurevych.2017.COLI}, a parallel strand of research tries to assess qualitative properties of arguments \cite{Habernal.Gurevych.2016.ACL,Stab.Gurevych.2017.EACL}. Yet the gap between theories and everyday argumentation, in understanding what `argument quality' actually is, remains an open research question \cite{Wachsmuth.et.al.2017.ACL,Habernal.Gurevych.2016.EMNLP}.

\parsum{Fallacies; it's hard to say whether an argument is good, but it might be easy to say if an argument is bad for some reason -- the whole field of fallacy research; but there is neither much empirical research nor data available}
Argumentation theories and critical thinking textbooks, however, offer an alternative view on quality of arguments, namely the notion of \emph{fallacies}: prototypical argument schemes or types that pretend to be correct and valid arguments but suffer logically, emotionally, or rhetorically \cite{Tindale.2007,Hamblin.1970}. Although this topic was first brought up by Aristotle already some 2,300 years ago, contemporary research on fallacies still does not provide a unifying view and clashes even in the fundamental questions \cite{Boudry.et.al.2015,Paglieri2016a}. Nevertheless, there seem to be several types of fallacies, such as \emph{argument ad hominem},\footnote{Attacking the opponent instead of her argument} various emotional \emph{appeals}, rhetorical moves of the \emph{red herring},\footnote{Distracting to irrelevant issues} or \emph{hasty generalization} that are, unfortunately, widely spread in our everyday argumentative discourse. Their powerful and sometimes detrimental impact was revealed in a few manual analyses \cite{Sahlane.2012,Nieminen.Mustonen.2014}. To the best of our knowledge, there is neither any NLP research dealing with fallacies, nor any resources that would allow for empirical investigation of that matter.

\parsum{Serious Games an promising alternative to dataset creation or annotation; so far only for very simple micro-tasks, works nice with pictures; gamification of more complex tasks an open question}
The lack of fallacy-annotated linguistic resources and thus the need for creating and labeling a new dataset from scratch
motivated us to investigate \emph{serious games} (also \emph{games with a purpose})---a scenario in which a task is \emph{gamified} and users (players) enjoy playing a game without thinking much of the burden of annotations \cite{vonAhn.Dabbish.2008,Mayer.et.al.2014.SeriousGames}. Serious games have been successful in NLP tasks that can be easily represented by images \cite{Jurgens.Navigli.2014,Kazemzadeh.et.al.2014.EMNLP} or that can be simplified to assessing a single word or a pair of propositions \cite{Neverilova.2014,Poesio.et.al.2013}. More complex tasks such as argument understanding, reasoning, or composing pose several design challenges centered around the key question: how to make data creation and annotation efforts fun and entertaining in the first place.

\parsum{We present Argotario: an online serious game for constructing dataset with fallacious argumentation}
To tackle this open research challenge, we created \emph{Argotario}---an online serious game for acquiring a dataset with fallacious argumentation.
The main research \textbf{contributions} and features of \emph{Argotario} include:
\begin{itemize}[noitemsep]
\item Gamification of the fallacy recognition task including player vs.\ player interaction
\item Learning by playing and educational aspects
\item Full in-game data creation and annotation, all data are under open license
\item Automatic gold label and quality estimation based solely on the crowd
\item Multilingual, platform independent, open-source, modular, with native look-and-feel on smartphones
\end{itemize}


\section{Background and Related Work}

\parsum{Very briefly fallacies, types, no single theoretical view}
Fallacies have been an active topic in argumentation theory research in the past several decades. While Aristotle's legacy was still noticeable in the twentieth century, a `fresh' look by \newcite{Hamblin.1970} showed that the concept of fallacies as arguments `that \emph{seem to be} valid but are \emph{not} so' deserves to be put under scrutiny.\footnote{\newcite{Hamblin.1970} criticized the `standard treatment' of fallacies widely present in contemporary textbooks as being `debased,' `worn-out', `dogmatic' and `without a connection to modern logic'.}
Theories about fallacies evolved into various categorizations and treatments, ranging from rather practical education-oriented approaches \citep{Tindale.2007,Schiappa.Nordin.2013} to rhetorical ones in informal logic \citep{Walton.1995} or pragma-dialectic \citep{vanEemeren.Grootendorst.1987}. For a historical overview of fallacies see, e.g., \citep{Hansen.2015}.

Surprisingly, the vast majority of current works on fallacies, and especially textbooks, present only toy examples that one is unlikely to encounter in real life \cite[p.~432]{Boudry.et.al.2015}. The distinction between fallacies and acceptable inference is fuzzy and theories do not offer any practical guidance: fully-fledged fallacies are harder to find in real life than is commonly assumed \cite{Boudry.et.al.2015}. To this account, analysis of fallacies in actual argumentative discourse has been rather limited in scope and size. \newcite{Nieminen.Mustonen.2014} examined fallacies found in articles supporting creationism.  \newcite{Sahlane.2012} manualy analysed fallacies in newswire editorials in major U.S. newspapers before invading Iraq in 2003. These two works rely on a list of several fallacy types, such as  \emph{ad hominem}, \emph{ad populum}, \emph{appeal to guilt}, \emph{slippery slope}, \emph{hasty generalization}, and few others. 

\parsum{Some serious games for NLP; their drawbacks: lack of interactivity (is it fun?) or simple tasks (symbols, not language); missing educational part (what I learn by playing the game)}
When scaling up annotations and resource acquisitions, serious games provide an alternative to paid crowdsourcing.
Recent successful applications include knowledge base extension \cite{Vannella.et.al.2014}, answering quizes related to medical topics \cite{Ipeirotis.Gabrilovich.2014}, word definition acquisition \cite{Parasca.et.al.2016}, or word sense labeling \cite{Venhuizen.et.al.2013}; where the latter one resembles a standard annotation task with bonus rewards rather than a traditional entertaining game.
\newcite{Niculae.Danescu-Niculescu-Mizil.2016.NAACL} built a game for guessing places given Google Street View images in order to collect data for investigating constructive discussions.
An important aspect of serious games for NLP is their benefit to the users other than getting the annotations done quickly: learning a language in \emph{Duolingo}\footnote{Although Duolingo presents itself as a learning tool, its incentives and competition features make it feel like accomplishing quests in a game.} has more added value than killing zombies (despite its obvious fun factor) in \emph{Infection} \cite{Vannella.et.al.2014}.


\section{Argotario: Overview}

\begin{figure*}
  \centering
  \subfloat[A single \emph{world} with the two first \emph{levels} finished, the third one about to be played, and other to be `explored'.]{\includegraphics[width=0.232\textwidth]{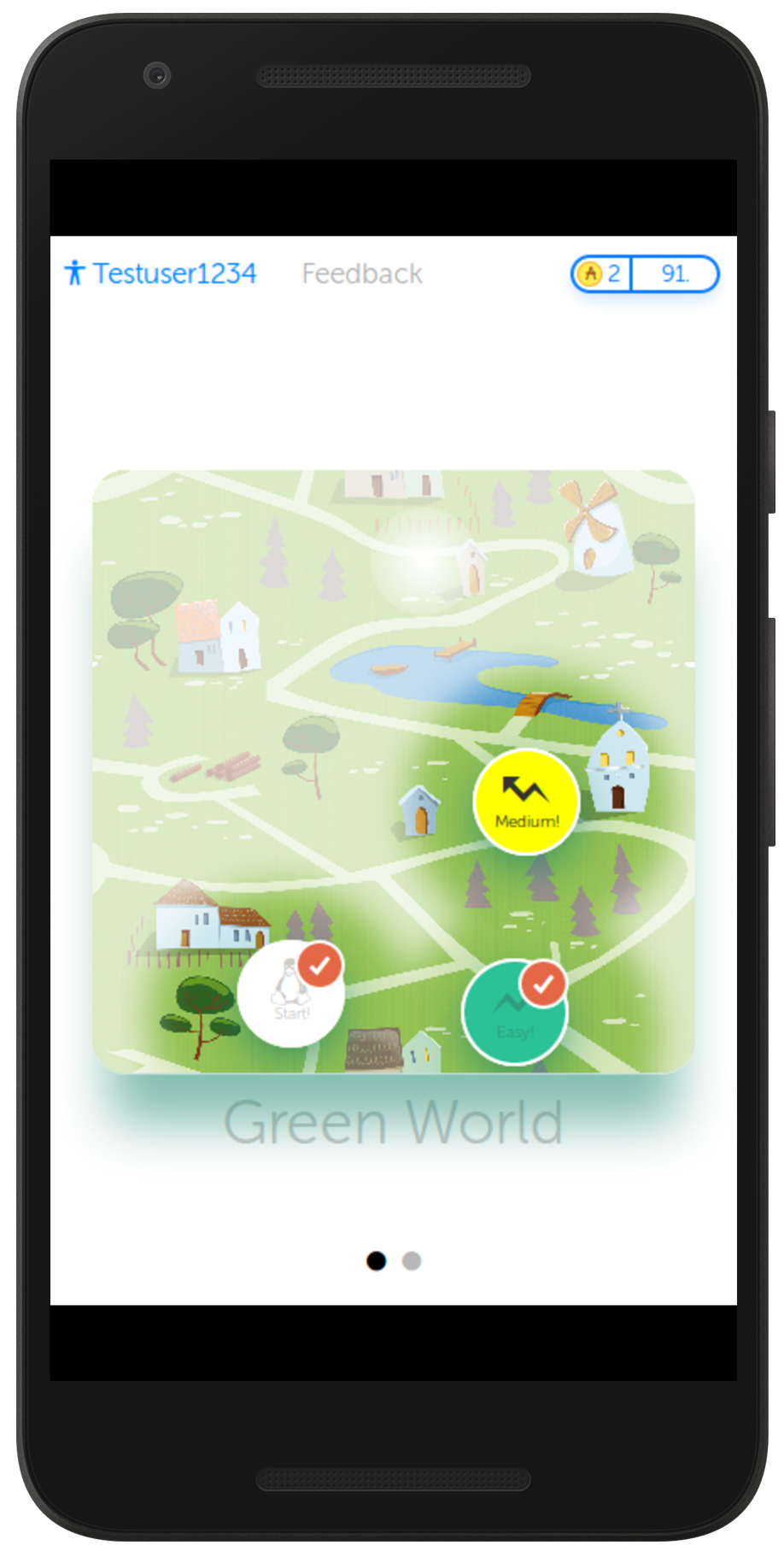}
  \label{fig:f1}}
  \hfill
  \subfloat[The recognize fallacy type \emph{round}.]{\includegraphics[width=0.229\textwidth]{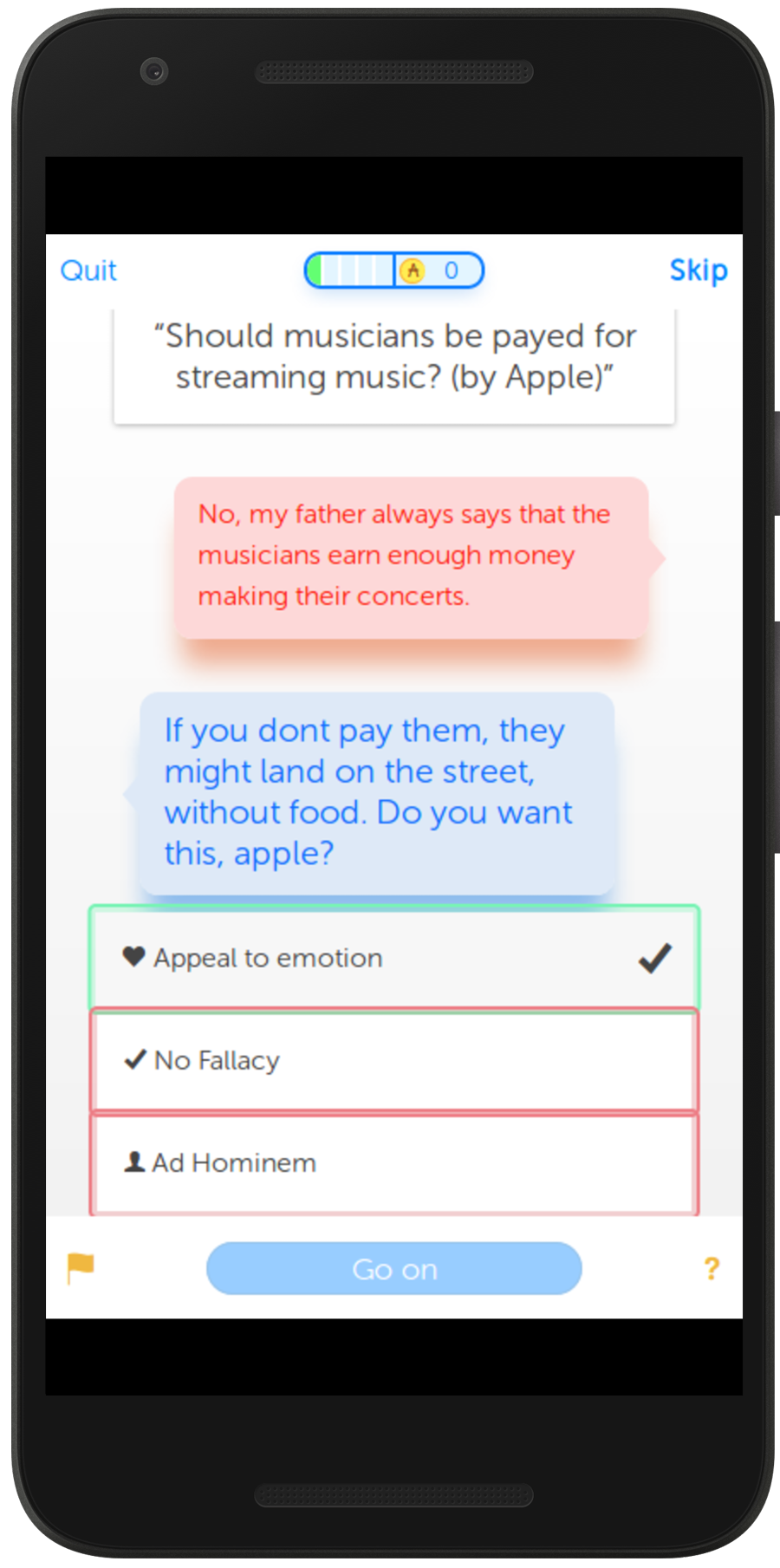}
  \label{fig:f3}}
  \hfill
  \subfloat[The \emph{player vs.\ player} level, now waiting for the opponent's turn.]{\includegraphics[width=0.23\textwidth]{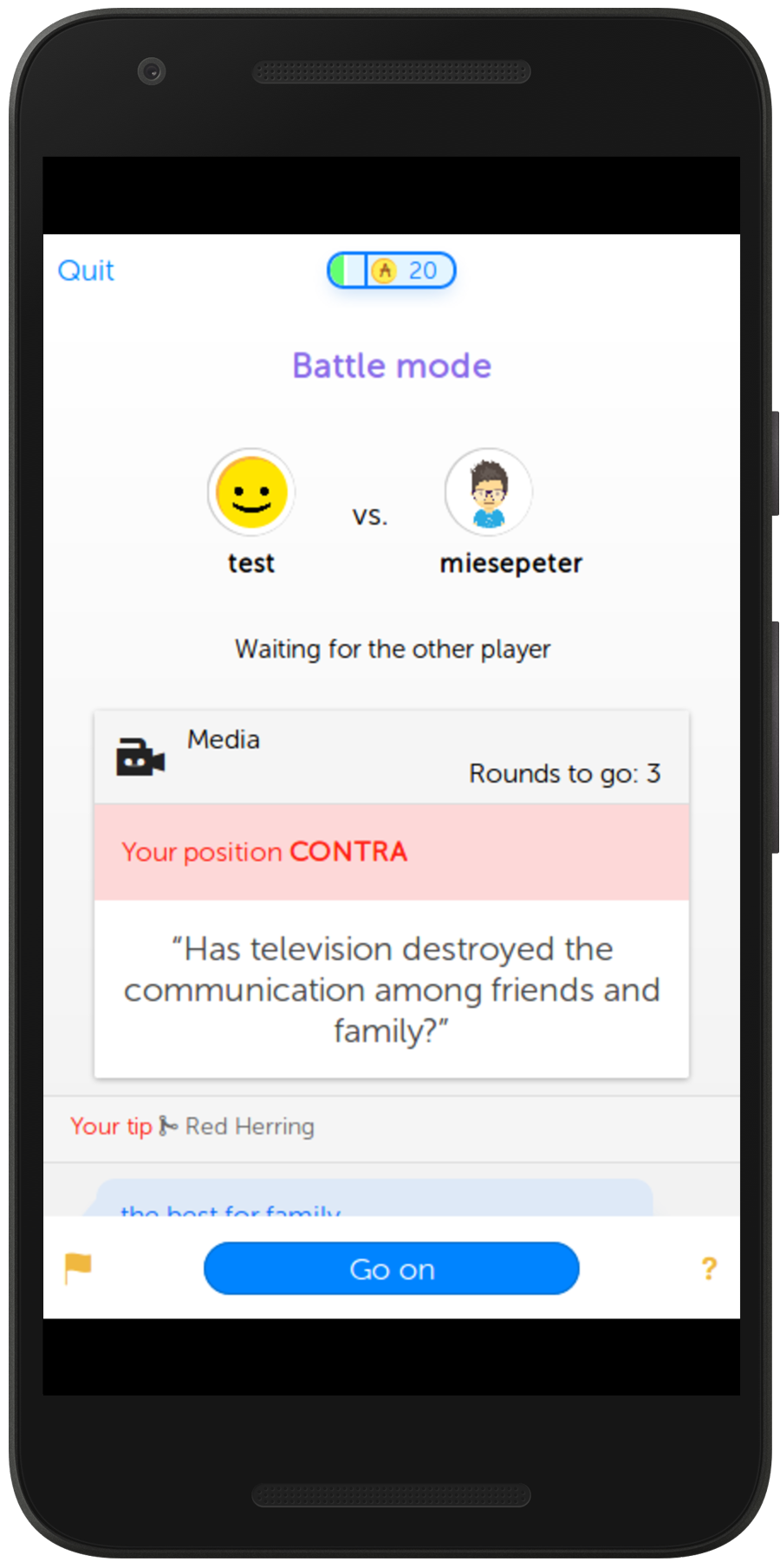}
  \label{fig:f4}}
  \hfill
  \subfloat[An example of \emph{hard feedback} in a fallacy recognition round.]{\includegraphics[width=0.231\textwidth]{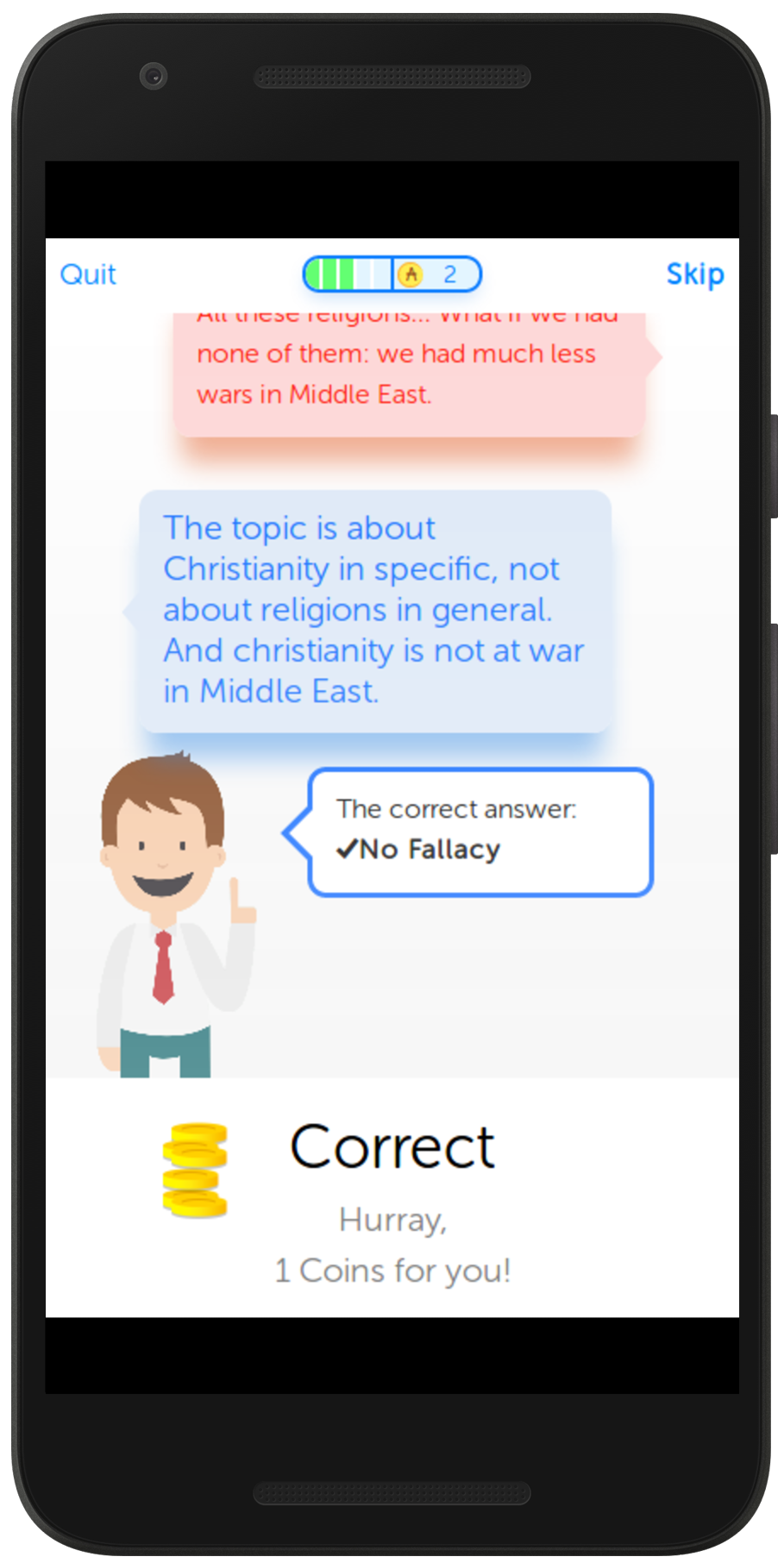}
  \label{fig:f2}}  
  \vspace{-0.5em}
  \caption{Screenshots of \emph{Argotario} taken in a smartphone emulator.}
  \vspace{-0.5em}
\end{figure*}

\paragraph{Architecture and Implementation}
\parsum{High-level overview: Online application, centralized, web based, native look-and-feel for smartphones yet runs in web-browsers}
\emph{Argotario} is a client-server Web-based application that runs in all modern browsers and seamlessly works on smartphones, providing an authentic look-and-feel. Its three-tier architecture consists of a backend MongoDB database, a Python server behind an Apache2 SSL proxy, and a Javascript client built on top of Ionic framework. \emph{Argotario} is \textbf{modular} as it allows developers to add new content (worlds, levels, rounds) as independent modules. The game workflow is \textbf{configurable} using JSON files, so it can be customized for evaluating new game scenarios. \textbf{Security} is ensured by a SSL certificate and securely hashing all passwords with salt. \textbf{Localization} utilizes the built-in capabilities of \texttt{ng-translate} so that all texts are stored externally in a JSON file and adding another language to the UI requires only manual translation of these texts.\footnote{Needless to say that providing an initial \emph{content} for a new language, such as a list of language-dependent topics, arguments, and fallacies, requires substantial manual work.} Currently, \emph{Argotario} is available in English and German.

\paragraph{Game Design}
\label{sec:game-design}

\parsum{User accounts; worlds; levels; rounds (round = microtask); coins as reward; scores}
We first present the abstract architecture; concrete examples follow in \S \ref{ref:gamifying}. According to \newcite[p.~50]{Salen.Zimmerman.2004}, a game is a system consisting of different types of interacting entities that have certain attributes. \emph{Argotario} follows this structure by a hierarchy of \textbf{worlds}, \textbf{levels}, and game \textbf{rounds} \cite{Hannemann.2015}.

A game \textbf{round} represents an atomic mini-game in which users take an action and are rewarded with points. Conceptually, each game round follows the same procedure: the users are first faced with \emph{game data}, which they need to interact with. Their response (a choice or free-text input) is then validated with respect to the current game round configuration, similar to form validation on web pages. If the game determines correctness of the response data, it rewards the user with a certain number of points.

A sequence of game rounds form a \textbf{level}. To complete a level, all game rounds must
be finished, independently of whether the user successfully fulfilled the respective task or not. Whereas game rounds can be re-used in different levels, each level is unique and can be individually designed to fit a certain purpose (i.e., only some types of fallacies are dealt with).

Finally, all levels reside in a \textbf{world} which is a wrapper for all included levels, visually resembling a treasure map (see Figure \ref{fig:f1}). Their look can be freely customized to be visually appealing and capture a certain atmosphere or theme. There are multiple worlds within the game next to each other.

\textbf{Users} are represented as small circular comic faces (avatars). The first user's \textbf{goal} is to finish all levels in all worlds. Initially, the game worlds are covered by a fog, which can be cleared by the user by completing levels. \textbf{Ranking} (score) is the second important game goal.
Repeating levels allows users to collect more points and hence improve their global rank.

\section{Gamifying Fallacy Recognition}
\label{ref:gamifying}

The backbone principle of \emph{Argotario} can be summarized as follows. First, since a fallacious argument is one `that seems to be valid but is not so' \cite{Hamblin.1970}, users must try to `fool' other users by \textbf{writing} a fallacious argument of a given type without being revealed that this is in fact a fallacy. By writing a fallacious argument on purpose with the aim to `disguise' it as a valid argument, users get sensitive to the very gist of fallacious argumentation (such as rhetorical strategies, linguistic devices, logic, etc.). Second, users learn to \textbf{recognize} fallacies in existing arguments---either by revealing the correct fallacy type or stating that the given argument is not fallacious---and get feedback about their `debunking' skills (see Figure \ref{fig:f3}).\footnote{All written texts and user input are licensed under CC-BY.}

In the serious-game terminology of \newcite[p.~61]{vonAhn.Dabbish.2008}, recognizing the correct fallacy type combines the \emph{inversion-problem game} (the guesser produces the input that was originally given to the describer`) and a modification of the \emph{output-agreement game} (the guesser has to produce the same output as the crowd; details will be discussed later in \S \ref{sec:gold-label-estimation}).

\paragraph{Fallacy Types}
We gathered an inventory of fallacy types suitable for our game scenarios. Given the breadth and variety of fallacy types \cite{Tindale.2007,Govier.2010}, we conducted several pilot studies to identify types that are (1) common in everyday argumentative discourse, (2) are distinguishable from each other, and (3) have increasing difficulty.\footnote{Details are out of scope of this demo paper and will be reported in future works. Briefly, we started our selection by taking the `usual suspects' from the \emph{`gang of eighteen'} \cite{Woods.2013}.} The fallacy type inventory in \emph{Argotario} currently contains \emph{ad hominem}, \emph{appeal to emotion}, \emph{red herring}, \emph{hasty generalization}, \emph{irrelevant authority}, and a non-fallacious argument \cite{Pollak.2016}. 

Players learn to recognize different fallacy types gradually, as they accomplish each level. After finishing the first world in which all fallacy types are mastered, users can engage in the \emph{player versus player} world. Here, a dialogue exchange about a given controversy requires users to write fallacious arguments (as in the previous world) and guess which fallacy was used by its opponent (thus getting points for correct answers; details about gold data estimation are explained in the next section). This level is asynchronous; when a user writes a new argument, his opponents gets notification about the turn change, so they do not have to play at the same time (see Figure \ref{fig:f4}).

\paragraph{Gold Label Estimation}
\label{sec:gold-label-estimation}

Because all content is created within the game by players with different abilities to write or comprehend argumentation, we treat the data as \emph{noisy} in the first place. First, spam can be reported in all rounds and is submitted to the admins to take action. Second, we rely on MACE  \cite{Hovy.et.al.2013} for gold label estimation which we seamlessly integrated to the backend. For example, if the user has to write an argument of a given fallacy type, we treat the type only as a single `vote' and require another four players to guess the correct type of this fallacy in other levels. Only arguments that receive at least five `votes' are fed into MACE to establish their gold label.

By utilizing crowd voting and spam reporting, we indirectly aim for high-quality labels. Predicting gold labels can be further parametrized by a \emph{threshold} in MACE, which then provides only gold label estimates for instances whose entropy is below the threshold \cite[p.~1125]{Hovy.et.al.2013}. However, a deep analysis of the data quality is on our current research agenda.



\paragraph{Feedback and Incentives}
\label{sec:feedback-and-incentives}

\emph{Argotario} provides two types of feedback: \emph{soft} and \emph{hard} one. For labeling arguments with yet unknown gold label, users get only one point without knowing whether their answer was right (\emph{soft feedback}). For arguments with already estimated gold labels, \emph{hard feedback} (see Figure \ref{fig:f2}) is given: if the user makes an error, she receives no reward. Apparently, hard feedback is better from the educational point of view as one knows immediately whether her answer was right or wrong; however, users do not know in advance whether a current assessment gives them a soft or hard feedback, so they are inherently encouraged to try their best.

We also built in several sorts of incentives to keep the player engaged. First, \emph{Argotario} shows the overall leaderboard as well as \emph{weekly} ranks to ensure newcomers have chances to succeed, see \cite{Ipeirotis.Gabrilovich.2014} for details. Players of the week are publicly shown and receive a small monetary prize. Second, debunking fallacious arguments to familiar topics is reportedly entertaining for players interested in rhetoric, argumentation, or public deliberation, according to user feedback obtained after few classroom runs.

\section{Benchmarking}
\label{sec:benchmarking}

\parsum{Several user studies; usability of the approach; benchmarking}
So far we tested \emph{Argotario} in serveral user studies and beta-testing sessions. The first study on early versions of \emph{Argotario} examined the effect of hard feedback and the lack thereof on overall users' engagement in the game. We found that the soft feedback has no significant negative impact on the users' experience\footnote{Two user groups (20 and 17 participants, respectively) with the same game configuration but with either only soft or hard feedback; final questionnaire with Likert-scale questions; Mann-Whitney-U non-parametric test.} \cite{Hannemann.2015}.

In a subsequent study, we benchmarked the player vs.\ player level using Amazon Mechanical Turk (AMT). We asked workers to play a specially configured version of \emph{Argotario} in order to `win' 20 points required for submitting the HIT. As the player vs.\ player round needs two dialogue turns of two users and thus two or more people actively participating over a longer period of time, we also implemented a naive \emph{bot} for this study.\footnote{We trained a fallacy classifier system on existing arguments in the database using a Convolutional Neural Network based on GloVe embeddings \cite{Pennington.2014} and Keras framework, so the \emph{bot} tried to recognize a fallacy in its opponent arguments during the player vs.\ player discussion. For generating an answer, it simply looked up an existing fallacy to the given topic. On the one hand, it disobeyed the discourse flow, as it obviously did not coherently respond to its opponent. On the other hand, it allowed us to deploy the game as a HIT on AMT and get a sufficient number of player vs.\ bot games in a short time.} At the same time, we promoted the game on social media and attracted some non-paid users. Using this process, we could quickly test the entire game mechanism with a larger crowd, identify potential drawbacks, and gather about 1,160 hand-written fallacious arguments. We also experimented with various price per HIT (\$1--\$2) with respect to average playing time. While the number of rejected low-quality HITs remained negligible for all configurations, we did not observe any correlation between HIT prices and playing times ($\approx$ 18--26 min). Our interpretation is that the HIT price for benchmarking studies should be fair and reflect the study time but does not influence the quality \cite{Pollak.2016}.

\section{Conclusions and Outlook}

\emph{Argotario} is a serious game that serves several purposes. First, it is a software tool for computational linguistics research, as it focuses on fallacies in argumentative discourse, an important part of qualitative criteria in computational argumentation.
Second, it is software supporting learning and education. Its main educational purpose is to raise awareness---not only that fallacies do exist but they might be easily overlooked and misused in everyday argumentation.
Finally, \emph{Argotario} is also a data-acquisition and annotation tool that applies successful techniques for quality estimation from crowd-sourcing approaches. All content is created by users within the game, as opposed to usual annotation tools.

In the long run, we expect that \emph{Argotario} provides a feasible method for data acquisition as compared to standard crowdsourcing. First, a purely monetary-driven perspective is not always the deciding factor of playing additional levels, as shown by \newcite{Eickhoff.et.al.2012}. Second, `experts' from the crowd motivated by the potential for achievement can help engage in participation \cite{Ipeirotis.Gabrilovich.2014}.

In the current version, \emph{Argotario} is still a proof of concept. Its capabilities need to be verified at a large scale in order to reveal patterns in the game dynamics with impact on the overall user experience and quality; these cannot be easily anticipated on small-scale benchmarks (\S\ref{sec:benchmarking}). In this regard, any manual intervention (such as spam removal) needs to be automated.

\emph{Argotario} is accessible at \url{www.argotario.net} along with tutorial videos and runs in any modern web-browser, preferably on smartphones. It is also open-source, source codes under ASL license are available at GitHub.\footnote{\url{https://github.com/UKPLab/argotario}}



\bibliography{bibliography}
\bibliographystyle{emnlp_natbib}

\end{document}